\documentclass[10pt,twocolumn,letterpaper]{article}

\usepackage{cvpr}
\usepackage{times}
\usepackage{epsfig}
\usepackage{graphicx}
\usepackage{amsmath}
\usepackage{amssymb}
\usepackage{multirow}
\usepackage{array}
\usepackage{comment}

\usepackage[breaklinks=true,bookmarks=false]{hyperref}

\cvprfinalcopy 

\def\cvprPaperID{3595} 

\begin{document}

\title{Object-aware Aggregation with Bidirectional Temporal Graph \\ for Video Captioning}

\author{Junchao Zhang and Yuxin Peng\thanks{Corresponding author.}\\
Institute of Computer Science and Technology,\\
Peking University, Beijing 100871, China\\
{\tt\small pengyuxin@pku.edu.cn}
}

\maketitle

\begin{abstract}
Video captioning aims to automatically generate natural language descriptions of video content, which has drawn a lot of attention recent years. Generating accurate and fine-grained captions needs to not only understand the global content of video, but also capture the detailed object information. Meanwhile, video representations have great impact on the quality of generated captions. Thus, it is important for video captioning to capture salient objects with their detailed temporal dynamics, and represent them using discriminative spatio-temporal representations. 
In this paper, we propose a new video captioning approach based on \textbf{object-aware aggregation with bidirectional temporal graph (OA-BTG)}, which captures detailed temporal dynamics for salient objects in video, and learns discriminative spatio-temporal representations by performing object-aware local feature aggregation on detected object regions.
The main novelties and advantages are: 
(1) \textbf{Bidirectional temporal graph}: 
A bidirectional temporal graph is constructed along and reversely along the temporal order, which provides complementary ways to capture the temporal trajectories for each salient object.   
(2) \textbf{Object-aware aggregation}:
Learnable VLAD (Vector of Locally Aggregated Descriptors) models are constructed on object temporal trajectories and global frame sequence, which performs object-aware aggregation to learn discriminative representations. A hierarchical attention mechanism is also developed to distinguish different contributions of multiple objects.
Experiments on two widely-used datasets demonstrate our OA-BTG achieves state-of-the-art performance in terms of BLEU@4, METEOR and CIDEr metrics.
\end{abstract}

\section{Introduction}

\begin{figure}[!t]
    \begin{center}\includegraphics[width=1\linewidth]{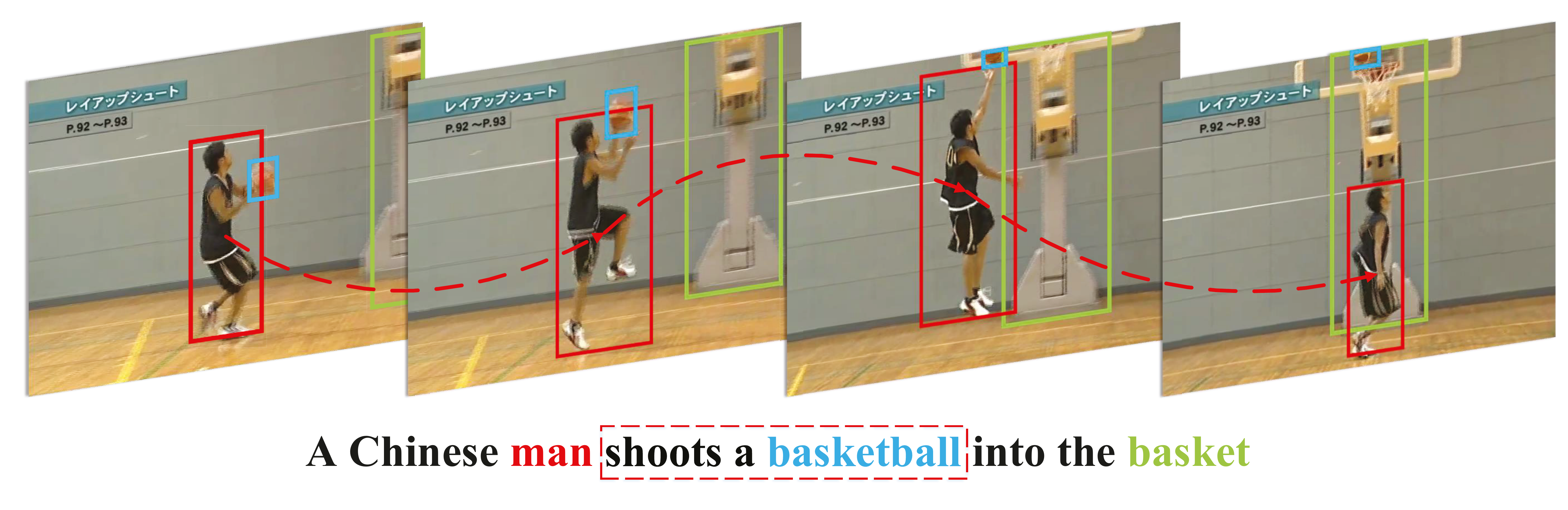}
    \caption{Illustration for the salient objects marked by the colored rectangle boxes and an example temporal trajectory indicated by dashed curve, which are important for discriminative video understanding to generate accurate captioning description.}
    \label{illustration}
    \end{center}
\end{figure}

As a task of generating natural language descriptions for video content automatically, video captioning takes a crucial step forward to the high-level video understanding and artificial intelligence. It supports various potential applications, such as human-robot interaction or assisting the visually-impaired. Recently, it has received increasing attention in both computer vision and artificial intelligence communities.


Previous works have explored to model the temporal information of video content by temporal attention mechanism \cite{yao2015describing,yang2017catching} or hierarchical encoder-decoder structures \cite{baraldi2017hierarchical,pan2016hierarchical}. However, they mainly work on the global frame or salient regions without discrimination on specific object instances, which cannot well capture the detailed temporal dynamics of each object.
While for obtaining the accurate captioning descriptions for complex video content, it plays a key but challenging role to capture the salient objects with their detailed temporal dynamics. As shown in Fig. \ref{illustration}, the reference captioning sentence ``A Chinese man shoots a basketball into the basket'' involves three salient objects in the example video, namely the boy, the basketball and the basket, which needs the object-aware video understanding. Besides, the reference sentence also describes the action that the boy is performing, ``shoots a basketball'', which needs to understand the detailed temporal dynamics of the boy and the basketball. 

In addition, video representations have great impact on the quality of generated captions. Therefore, how to describe the video content using discriminative spatio-temporal representations is also important for video captioning. Many works directly extract global features on video frames from the fully-connected layer or global pooling layer in CNN, which may lose much fine spatial information.
NetVLAD \cite{arandjelovic2016netvlad} shows its local information encoding ability by embedding a trainable VLAD (vector of locally aggregated descriptors) encoding model into the CNN, which aggregates the local features to encode local spatial information. Following it, SeqVLAD \cite{xu2018sequential} is proposed recently to combine the trainable VLAD encoding model with the sequence learning process, which explores both the local spatial information and temporal information of video. However, the above methods do not consider the object-specific information, thus cannot distinguish specific fine spatio-temporal information corresponding to the specific object instance. 

For addressing the above two problems, in this paper, we propose a novel video captioning approach based on \emph{object-aware aggregation with bidirectional temporal graph (OA-BTG)}, which captures detailed temporal dynamics for the salient objects in video via a bidirectional temporal graph, and learns discriminative spatio-temporal video representations by performing object-aware local feature aggregation on object regions. 
Its main novelties and advantages are:
\begin{itemize}
\item
\textbf{Bidirectional temporal graph}: 
The bidirectional temporal graph is constructed on salient objects and global frames to capture the detailed temporal dynamics in video. The bidirectional temporal graph includes a forward graph along the temporal order and a backward graph reversely along the temporal order, which provide different ways to construct the temporal trajectories with complementary information for each salient object instance. In such way, detailed temporal dynamics for objects and global context are captured to generate accurate and fine-grained captions. 
\item
\textbf{Object-aware aggregation}: 
For encoding the fine spatio-temporal information, we construct learnable VLAD models on object temporal trajectories and global frame temporal sequences, which perform object-aware aggregation for each salient object instance as well as the global frame to learn discriminative representations. We also utilize a hierarchical attention mechanism to distinguish different contributions of different object instances. In such way, we learn the discriminative spatio-temporal video representations for boosting the captioning performance. 
\end{itemize}
We conduct experiments on two widely-used datasets MSVD and MSR-VTT, which demonstrate that our proposed OA-BTG approach achieves the state-of-the-art performance in terms of BLEU@4, METEOR and CIDEr metrics for video captioning.


\section{Related Works}

In the early stage, video captioning methods are mainly template-based language models \cite{guadarrama2013youtube2text,rohrbach2013translating,krishnamoorthy2013generating}. These methods follow a bottom-up paradigm, which first predicts semantic concepts or words, like objects, scenes and activities, and then generates sentences according to pre-defined language templates. These methods heavily rely on the template definition and the predicted video concepts, which limits the diversities of generated sentences. 
Recently, inspired by the development of deep learning and neural machine translation (NMT) \cite{bahdanau2014neural}, many sequence learning based models \cite{pan2016hierarchical,venugopalan2015sequence,wang2018m3,ZhangP19Hierachical} are proposed to address video captioning problem. Regarding video captioning as a ``translating'' process, these methods construct the encoder-decoder structures to directly generate sentences from the video content.

Venugopalan et al. \cite{venugopalan2014translating} make the early attempt to generate video descriptions using encoder-decoder structure, but they simply apply mean pooling over individual frame features to obtain video representation, which ignores the temporal information of ordered video frames. For addressing this issue, the following works \cite{yao2015describing, pan2016hierarchical, venugopalan2015sequence} make advances by using temporal attention mechanism, as well as taking LSTM-based encoders to learn long-term temporal structures.
Yang et al. \cite{yang2017catching} achieve the progress by further considering the different characteristics of video frames. They propose to adaptively capture the regions-of-interests in each frame, then learn discriminative features based on these regions-of-interests for better video captioning. Xu et al. \cite{xu2018sequential} propose the SeqVLAD method, which performs feature aggregation on frame features to exploit fine spatial information in video content.
 
However, these methods mainly work on the global frame or salient regions without discrimination on specific object instances, which cannot well capture the temporal evolution of each object in video. In this work, we propose the OA-BTG approach, which constructs bidirectional temporal graph on the objects across video frames to captures their temporal trajectories. In addition, we also perform representation learning on the temporal trajectories, which exploits the object-awareness to boost video captioning.

There are also some works \cite{wang2018m3,chen2017video,xu2017learning,hori2017attention,WangWW18} that exploit multi-modal features for video captioning.
Besides frame features extracted by popular 2D CNNs, they also exploit motion features extracted by C3D \cite{tran2015learning} or audio features \cite{xu2013feature}, where they mine the complementarities among multi-modal information to boost the video captioning performance. Different from them, our OA-BTG approach takes only visual features, which mainly focuses on capturing detailed temporal evolutions of objects by bidirectional temporal graph and learning discriminative features through object-aware feature aggregation.

\section{Object-aware Aggregation with Bidirectional Temporal Graph}
\begin{figure*}[!t]
    \begin{center}\includegraphics[width=1.0\linewidth]{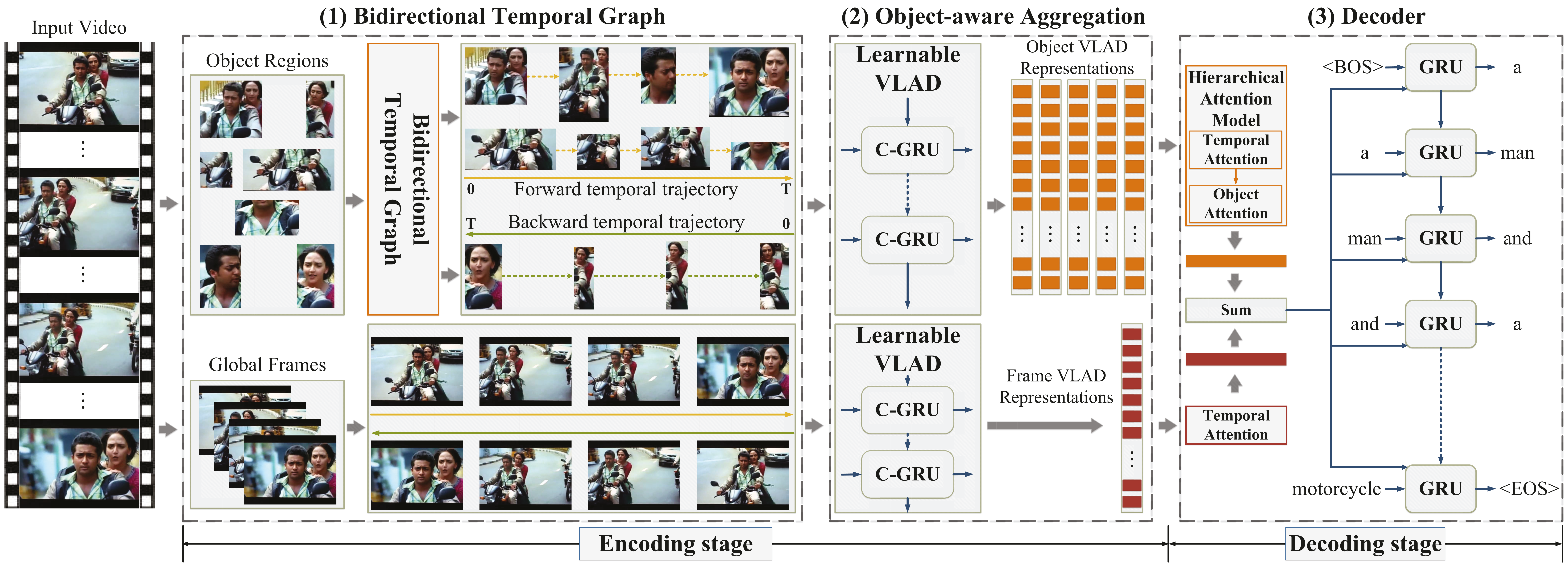}
    \caption{Overview of the proposed OA-BTG approach.}
    \label{framework}
    \end{center}
\end{figure*}

In this section, we present the proposed video captioning approach based on \emph{object-aware aggregation with bidirectional temporal graph (OA-BTG)} in detail, which follows the encoder-decoder framework.
As shown in Figure \ref{framework}, our OA-BTG consists of three components.
\textbf{(1) Bidirectional Temporal Graph}: For the input video, we first extract frames and multiple object regions. Then we construct bidirectional temporal graph to capture detailed temporal dynamics along and reversely along the temporal order.
\textbf{(2) Object-aware Aggregation}: Based on the bidirectional temporal graph, we perform object-aware aggregation to aggregate the local features of object regions and global frames into discriminative VLAD representations using learnable VLAD models.
\textbf{(3) Decoder}: Above two components form the encoding stage. While in the decoding stage, the learned object and frame VLAD representations are integrated and fed into the GRU units to generate descriptions. Especially, hierarchical attention is applied to distinguish the different contributions of multiple objects.
In the following subsections, we will introduce the bidirectional temporal graph, object-aware aggregation and decoder respectively. 

\subsection{Bidirectional Temporal Graph}
The \emph {bidirectional temporal graph (BTG)} is constructed based on detected object regions and global frames. For object regions, bidirectional temporal graph is constructed to group the same object instances or similar regions together along and reversely along the temporal order. It obtains the forward and backward temporal trajectories of each detected object instance, thus capture the detailed temporal dynamics of salient objects, which are important for generating accurate and fine-grained language descriptions.
For the global frames, we organize them into sequences along and reversely along the temporal order to capture the forward and backward temporal dynamics for global context.

The bidirectional temporal graph is constructed according to the similarities among object regions in different frames, including a forward graph and a backward graph to obtain the object trajectories along and reversely along the temporal order respectively. Specifically, for each frame $v_t$ in the input video $V$, we extract $N$ object regions, $R^{(t)}=\{r^{(t)}_i\}$, where $i=1,2,\cdots,N$, $t=1,2,\cdots, T$, and $T$ is the number of sampled frames of $V$.
For constructing the \emph{forward graph}, we take the object regions in frame $v_1$ as anchors to compute the similarities with object regions in all other frames, where the similarity score is defined to indicate whether the two object regions belong to the same object instance.
With jointly considering the appearance information and relative spatial location information between two object regions $r_i$ and $r_j$, we define their similarity score $s(i,j)$ as follows:
\begin{equation}
s(i,j) = (s_{app}(i,j)+s_{iou}(i,j)+s_{area}(i,j))/3
\label{similarity}
\end{equation}

As for three terms in above equation, $s_{app}$ indicates similarity on visual appearance between $r_i$ and $r_j$, which is computed according to the Euclidean distance of their visual features:
\begin{equation}
s_{app}(i,j) = exp\left(-\frac{L_2(g_i, g_j)}{max_{p,q}(L_2(g_p,g_q))}\right)
\end{equation}
where $L_2$ denotes the Euclidean distance, and $max_{p,q}(L_2(g_p,g_q))$ computes the maximal Euclidean distance of all object region pairs between the corresponding two frames, which is utilized as a normalization factor. $g$ indicates the extracted visual feature for object region using pretrained CNN model, taking the cropped object region image as input.
$s_{iou}$ and $s_{area}$ indicate the rates of overlap area and area sizes between $r_i$ and $r_j$, respectively, which are computed as follows:
\begin{equation}
s_{iou}(i,j) = \frac{area(r_i)\cap area(r_j)}{area(r_i)\cup area(r_j)}
\end{equation}
\begin{equation}
s_{area}(i,j) = exp\left(-\left|\frac{min(A_i, A_j)}{max(A_i, A_j)}-1\right|\right)
\end{equation}
where $area$ denotes the spatial area of the object region and $A$ mean its area size.

The construction of the \emph{backward graph} is similar to the forward graph, while the anchors to compute the similarity scores are composed of the object regions in frame $v_T$. Then the forward graph and the backward graph are combined to compose the bidirectional temporal graph.

Aiming to group the object regions in different frames but belonging to the same object instance together, we compare object regions in all the other frames with the anchor object regions, and then align them to the anchor object regions with a nearest neighbor (NN) strategy according to the bidirectional temporal graph. Specifically, for the object region $r^{(1)}_i$ in anchor frame $v_1$ and $N$ object regions $R^{(t)}=\{r^{(t)}_j\}$ in frame $v_t$, $t=2,\cdots,T$, the object region $\mathop{argmax}_{r^{(t)}_j}(s(i,j))$ is aligned to the object region $r^{(1)}_i$, which means they are considered to belong to the same object instance. We also align object regions in other frames to the anchor object regions in frame $v_T$ using the same NN strategy.

After above alignment process, we have obtained $2N$ groups of aligned object regions. Then, $N$ groups on the forward graph are organized along the temporal order to obtain the forward temporal trajectories of detected object instances, while the other $N$ groups on the backward graph are organized reversely along the temporal order to obtain the backward temporal trajectories. The forward and backward temporal trajectories are complement on capturing the detailed temporal dynamics of salient objects in video for the following two aspects: (1) Organizing the temporal sequence along and reversely along the temporal order provides two different ways to represent the temporal dynamics of video content, and thus provides complementary information. (2) It usually cannot obtain the good temporal trajectories for all the salient objects only on the forward or backward graph, since not all the objects occur throughout the whole video. Thus, we resort to both forward and backward temporal trajectories on the bidirectional temporal graph for capturing the temporal dynamics better. 

We denote the forward and backward temporal trajectories as $O^f_i=\{o^f_{it}\}$ and $O^b_i=\{o^b_{it}\}$, $t=1,\cdots,T$, $i=1,\cdots,N$, respectively. Then, we aslo organize the global frames along and reversely along the temporal order as $V^f=\{v^f_t\}$ and $V^b=\{v^b_t\}$, $t=1,\cdots,T$, to capture comprehensive temporal dynamics on global context.

\subsection{Object-aware Aggregation}
For \emph{object-aware aggregation (OA)}, we devise two learnable VLAD models to learn the spatio-temporal correlations for object region sequences and global frame sequence, as well as aggregate the local features of object regions and frames into discriminative VLAD representations.

Local features are first extracted for global frames and the detected objects of input video $V$. We feed the global frames and cropped object region images into the pretrained CNN model and take the feature maps from the convolutional layer as local features. Each feature map has the size as $H \times W \times D$, where $H$, $W$ and $D$ mean the height, width and the number of channels. We organize the local features of object regions and global frames according to the forward and backward temporal sequences obtained based on the BTG, and denote them as $X^{of}_i=\{x^{of}_{it}\}$ and $X^{ob}_i=\{x^{ob}_{it}\}$ for object sequences, as well as $X^f=\{x^f_t\}$ and $X^b=\{x^b_t\}$ for frame sequences.

Inspired by NetVLAD \cite{arandjelovic2016netvlad} and SeqVLAD \cite{xu2018sequential}, we utilize a convolutional gated recurrent unit (C-GRU) architecture to construct the learnable VLAD model, where the C-GRU aims to learn the soft assignments for VLAD encodings.

Taking the learnable VLAD model on forward temporal sequences of object regions as an example, the C-GRU takes the local feature $x^{of}_{t}$ (here the subscript $i$ is omitted for simplicity) at time $t$ and the hidden state $a_{t-1}$ at time $t-1$ as inputs, and then updates its hidden state $a_{t}$ as follows:
\begin{equation}
z_t = \sigma(W_z * x^{of}_{t} + U_z * a_{t-1})
\end{equation}
\begin{equation}
r_t = \sigma(W_r * x^{of}_{t} + U_r * a_{t-1})
\end{equation}
\begin{equation}
\widetilde{a}_t = tanh(W_a * x^{of}_{t} + U_a * (r_t \odot a_{t-1}))
\end{equation}
\begin{equation}
a_t = (1 - z_t) \odot a_{t-1} + z_t \odot \widetilde{a}_t
\end{equation}
where $W_z, W_r, W_a$ and $U_z, U_r, U_a$ denote the 2D-convolutional kernels. Noted that all $N$ groups of object region sequences share the same C-GRU parameters. $*$ denotes the convolution operation, $\sigma$ denotes the sigmoid activation function, and $\odot$ denotes the element-wise multiplication. The output hidden state $a_t \in \mathbb{R}^{H\times W\times K}$ denotes the learned assignments between local features $X^{of}$ and $K$ cluster centers, which are also learnable and introduced below.

VLAD is a feature encoding method that learns $K$ cluster centers as visual words, denoted as $C = \{c_k\}$, $k=1,\cdots,K$, and then maps each local feature to the nearest $c_k \in \mathbb{R}^D$. Its key idea is to accumulate the differences between local features and the corresponding cluster center. Inspired by SeqVLAD \cite{xu2018sequential}, we set $K$ cluster centers as learnable parameters and adopt ``soft assignment'' strategy that assign the local features to the cluster centers according to the learned assignment parameters $a_t$:
\begin{equation}
vl^{of}_t = \sum_{h=1}^H \sum_{w=1}^W a_t(h, w, k)(x^{of}_t-c_k)
\end{equation}
Then we obtain the VLAD representations for forward temporal sequences of object regions as $VL^{of}_{i}=\{vl^{of}_{it}\}$, $t=1,\cdots,T$, and $i=1,\cdots,N$. Similarly, the VLAD representations for backward sequences of object regions as well as two temporal sequences of global frames can be obtained as $VL^{ob}_{i}=\{vl^{ob}_{it}\}$, $VL^{f}=\{vl^{f}_{t}\}$ and $VL^{b}=\{vl^{b}_{t}\}$.

\subsection{Decoder}
In the decoding stage, we process the forward and backward temporal sequences respectively. Taking the forward temporal sequences as example, the decoder is constructed by GRU units with attention mechanism, which utilizes VLAD representations of objects $\{VL^{of}_{i}\}$ and frames $VL^{f}$ to generate words for captioning.

As shown in Fig. \ref{framework}, the attention model for object VLAD representations has a hierarchical structure including temporal attention and object attention. The temporal attention is applied to highlight object regions at discriminative time steps when merging $T$ object VLAD representations into one representation, while the object attention is designed to distinguish the different contributions of $N$ different object instances. The temporal attention mechanism is formulated as follows:
\begin{equation}
e_{lt} = w_{att}^T tanh(W_{att}h_{l-1} + U_{att}vl^{of}_{it} + b_{att})
\end{equation}
where 
$w_{att}, W_{att}, U_{att}$ and $b_{att}$ are parameters. $e_{lt}$ computes the relevant score between the visual feature $vl^{of}_{it}$ and the hidden state $h_{l-1}$ of GRU decoder at time $l-1$, $l \leq L$ indicates the time step at decoding stage. Then the relevance scores are normalized as attention scores:
\begin{equation}
\beta_{lt} = exp(e_{lt}) / \sum_{n=1}^T exp(e_{ln})
\end{equation}
Then $T$ object visual features are merged according to above attention scores:
\begin{equation}
\phi^{of}_{li} = \sum_{t=1}^T \beta_{lt}vl^{of}_{it}
\end{equation}

The object attention takes the same mechanism as temporal attention, which is applied on $\{\phi^{of}_{li}\}$, $i=1,\cdots,N$ to distinguish the different contributions of $N$ different object instances:
\begin{equation}
e^o_{li} = w_{att}^{oT} tanh(W^o_{att}h_{l-1} + U^o_{att}\phi^{of}_{li} + b^o_{att})
\end{equation}
\begin{equation}
\beta^o_{li} = exp(e^o_{li}) / \sum_{n=1}^N exp(e^o_{li})
\end{equation}
\begin{equation}
\phi^{of}_{l} = \sum_{i=1}^N \beta^o_{li}\phi^{of}_{li}
\end{equation}
where $w_{att}^{o}, W^o_{att}, U^o_{att}$ and $b^o_{att}$ are parameters, and $\phi^{of}_{l}$ denotes the discriminative spatio-temporal feature that indicates the object information.

For $T$ frame VLAD representations $\{vl^{f}_{t}\}$, the temporal attention mechanism is applied on them to obtain the discriminative spatio-temporal feature $\phi^{f}_{l}$ that indicates the global context information.

At time $l$, the obtained features $\phi^{of}_{l}$ and $\phi^{f}_{l}$ are fed into the GRU unit to update the hidden state and generate the word:
\begin{equation}
z^d_l = \sigma(W_{vz} \phi^{f}_{l} + W_{oz} \phi^{of}_{l} + W_{dz} x^w_l + U_{dz} h_{l-1})
\end{equation}
\begin{equation}
r^d_l = \sigma(W_{vr} \phi^{f}_{l} + W_{or} \phi^{of}_{l} + W_{dr} x^w_l + U_{dr} h_{l-1})
\end{equation}
\begin{equation}
\widetilde{h}_l = tanh(W_{vh} \phi^{f}_{l} + W_{oh} \phi^{of}_{l} + U_{dh}(r^d_l \odot h_{l-1}))
\end{equation}
\begin{equation}
h_{l} = (1 - z^d_l) \odot h_{l-1} + z^d_l \odot \widetilde{h}_l
\end{equation}
where $\sigma$ denotes the sigmoid function and $x^w_l$ denotes the word embedding for the input word in time $l$. $W_{v*}, W_{o*}, W_{d*}$ and $U_{d*}$ denote the parameters to learn.

After obtaining the hidden state $h_{l}$, we apply a linear layer and a softmax layer to compute the probability distribution over all the vocabulary words. In the training stage, we utilize the cross-entropy loss to optimize all the learnable parameters. While in the testing stage, we take beam search method to generate the captioning descriptions.

We take a simple but effective way to exploit the complementarity of the forward and backward temporal sequences (corresponding to the forward and backward graphs respectively). In each time step, we fuse the obtained predicted scores of words based on forward and backward graphs, and then the word is generated according to the fused scores. In such way, we mine the complementary between the forward and backward temporal sequences to boost the video captioning performance.

\section{Experiments}

\subsection{Datasets and Evaluation Metrics}
\subsubsection{Datasets.} 
We evaluate our proposed OA-BTG approach on two widely-used datasets, including Microsoft Video Description Corpus (MSVD) \cite{chen2011collecting} and Microsoft Research-Video to Text (MSR-VTT) \cite{xu2016msr}.

\textbf{MSVD} is an open-domain dataset for video captioning that covers various topics including sports, animals and music. It contains 1,970 video clips from Youtube and collects multi-lingual descriptions by Amzon Mechanical Turk (AMT). There are totally about 8,000 English descriptions with roughly 40 descriptions per video. 
Following \cite{yang2018video,xu2017learning}, we only consider the English descriptions, taking 1,200, 100, 670 clips for training, validation, testing.


\textbf{MSR-VTT} is a large-scale benchmark used in the video-to-language challenge\footnote{We adopt the dataset of 2016's challenge, and the corresponding competition results can be found in http://ms-multimedia-challenge.com/2016/leaderboard.}. It contains 10,000 video clips with 200,000 clip-sentence pairs in total, and covers comprehensive video categories, diverse video content as well as language descriptions. There are totally 20 categories, such as music, sports, movie, etc. The descriptions are also collected by AMT, and each video clip is annotated with 20 natural language sentences. Following the splits in \cite{xu2016msr}, there are 6,513 clips for training, 497 clips for validation, and 2,990 clips for testing.

\subsubsection{Evaluation Metrics.}
For quantitative evaluation of our proposed approach, we adopt the following common metrics in our experiments: \textbf{BLEU@4} \cite{papineni2002bleu}, \textbf{METEOR} \cite{banerjee2005meteor}, and \textbf{CIDEr} \cite{vedantam2015cider}. BLEU@4 measures the fraction of $n$-grams (here $n=4$) between generated sentence and ground-truth descriptions. METEOR measures uni-gram precision and recall between generated sentence and ground-truth references, extending exact word matching to including similar words. CIDEr is a voting-based metric, which to some extent is robust to incorrect ground-truth descriptions.
Following \cite{venugopalan2015sequence,pan2016hierarchical,yang2017catching}, all the metrics are computed by using the Microsoft COCO evaluation server \cite{chen2015microsoft}.

\subsection{Experimental Settings}
\subsubsection{Video and sentence preprocessing.} 

We sample $40$ ($T=40$) frames for each input video and extract $5$ ($N=5$) objet regions for each frame empirically. We utilize Mask R-CNN \cite{he2017mask} to detect objects, which is based on ResNet-101 \cite{he2016deep} and pre-trained on Microsoft COCO dataset \cite{lin2014microsoft}. All the object regions are cropped into images and fed into ResNet-200 to extract local features. The global frames are also fed into ResNet-200 to obtain local features. We take the output of $res5c$ layer in ResNet-200 as local feature map with the size of $7 \times 7\times 2048$. 

All the reference captioning sentences are tokenized and converted to lower case. After removing the punctuations, we collect $12,593$ word tokens for MSVD dataset and $27,891$ word tokens for MSR-VTT dataset.

\subsubsection{Training details.} For the training video/sentence pairs, we filter out the sentences with more than 16 words, and adopt zero padding strategy to complement the sentences that has less than 16 words. 
During training, begin-of-sentence $<$BOS$>$ tag and end-of-sentence $<$EOS$>$ tag are added at the beginning and end of each sentence. Unseen words in the vocabulary will be set to $<$UNK$>$ tags. Each word is encoded as a one-hot vector.
The hidden units of encoder and decoder are set as $512$. The word embedding size and attention size are set as $512$ and $100$ respectively. For the trainable VLAD models, we set the cluster center number $K$ as $64$ for MSVD dataset and $128$ for MSR-VTT dataset. The reason is that MSR-VTT is a large-scale dataset with diverse video content, thus a larger number of cluster centers are necessary to fully represent the video content.

During training stage, all the parameters are randomly initialized, and we utilize Adam algorithm to optimize captioning model. The learning rate is fixed to be $1 \times 10^{-4}$, and the training batch size is 16. 
Dropout is applied on the output of decoder GRU with the rate of $0.5$ to avoid overfitting. We also apply gradient clip of $[-10,10]$ to prevent gradient explosion.
In testing stage, we adopt beam search to generate descriptions, where beam size is set as $5$.

\begin{table}[ht]
  \centering
  \caption{Comparisons with state-of-the-art methods on MSVD dataset. 
  All the results are reported as percentage (\%). 
  }
  \scalebox{0.95}{
  \begin{tabular} {c|c|c|c}
    \hline
    Methods & BLEU@4 & METEOR & CIDEr\\
    \hline
    \hline
    \textbf{Our OA-BTG}  & \textbf{56.9} & \textbf{36.2} & \textbf{90.6} \\
    \hline
    \hline
    SeqVLAD \cite{xu2018sequential}                        & 51.0 & 35.2 & 86.0 \\
    LSTM-GAN \cite{yang2018video}                                   & 42.9 & 30.4 & -    \\
    MS-RNN \cite{song2018deterministic}                    & 53.3 & 33.8 & 74.8 \\
    MCNN+MCF \cite{wu2018multi}                                     & 46.5 & 33.7 & 75.5 \\
    RecNet \cite{wang2018reconstruction}                            & 52.3 & 34.1 & 80.3 \\
    TSA-ED \cite{wu2018interpretable}                      & 51.7 & 34.0 & 74.9 \\
    aLSTMs \cite{gao2017video}                                 & 50.8 & 33.3 & 74.8 \\
    STAT \cite{tu2017video}                                  & 51.1 & 32.7 & 67.5 \\
    MA-LSTM \cite{xu2017learning}                                 & 52.3 & 33.6 & 70.4 \\
    DMRM \cite{yang2017catching}                                    & 51.1 & 33.6 & 74.8 \\
    hLSTMat \cite{song2017hierarchical}                     & 53.0 & 33.6 & 73.8 \\
    mGRU \cite{zhu2017bidirectional}                       & 53.8 & 34.5 & 81.2 \\
    TDDF \cite{zhang122017task}                                   & 45.8 & 33.3 & 73.0 \\
    \hline
  \end{tabular}
  }
  \label{comparisons_msvd}
\end{table}

\begin{table}[ht]
  \centering
  \caption{Comparisons with state-of-the-art methods on MSR-VTT dataset. 
  All the results are reported as percentage (\%). 
  }
  \scalebox{0.95}{
  \begin{tabular} {c|c|c|c}
    \hline
    Methods & BLEU@4 & METEOR & CIDEr\\
    \hline
    \hline
    \textbf{Our OA-BTG}  & \textbf{41.4} & \textbf{28.2} & \textbf{46.9} \\ 

    \hline
    \hline
    LSTM-GAN \cite{yang2018video}                                   & 36.0 & 26.1 & -    \\
    MS-RNN \cite{song2018deterministic}                    & 39.8 & 26.1 & 40.9 \\
    MCNN+MCF \cite{wu2018multi}                            & 38.1 & 27.2 & 42.1 \\
    M$^3$ \cite{wang2018m3}                                       & 38.1 & 26.6 & -    \\ 
    RecNet \cite{wang2018reconstruction}                            & 39.1 & 26.6 & 42.7 \\
    aLSTMs \cite{gao2017video}                                 & 38.0 & 26.1 & 43.2 \\
    STAT \cite{tu2017video}                                  & 37.4 & 26.6 & 41.5 \\
    MA-LSTM \cite{xu2017learning}                                 & 36.5 & 26.5 & 41.0 \\
    hLSTMat \cite{song2017hierarchical}                    & 38.3 & 26.3 & -    \\
    TDDF \cite{zhang122017task}                                   & 37.3 & 27.8 & 43.8 \\
    \hline
  \end{tabular}
  }
  \label{comparisons_msrvtt}
\end{table}

\subsection{Comparisons with State-of-the-art Methods}

\begin{table*}[!ht]
  \centering
  \caption{Effectiveness of different components in our OA-BTG approach on MSVD and MSR-VTT datasets. All the results are reported as percentage (\%). 
  }
  \scalebox{0.95}{
  \begin{tabular} {c|c|c|c|c|c|c}
    \hline
    \multirow{2}{*}{Methods} & \multicolumn{3}{c|}{MSVD} & \multicolumn{3}{c}{MSR-VTT} \\

    \cline{2-7}& BLEU@4 & METEOR & CIDEr & BLEU@4 & METEOR & CIDEr \\
    \hline  
    BASELINE                 & 52.7 & 34.1 & 83.7        & 39.6 & 26.3 & 42.3 \\
    \hline                         
    Our OA with Forward TG       & 54.0 & 34.7 & 84.9        & 40.8 & 26.9 & 45.1 \\
    Our OA with Backward TG      & 53.3 & 35.4 & 85.4        & 40.8 & 27.3 & 45.3 \\
    \hline
    \textbf{Our OA-BTG}   & \textbf{56.9} & \textbf{36.2} & \textbf{90.6} & \textbf{41.4} & \textbf{28.2} & \textbf{46.9} \\
    \hline
  \end{tabular}
  }
  \label{baselines}
\end{table*}

Tables \ref{comparisons_msvd} and \ref{comparisons_msrvtt} show comparative results between OA-BTG and the state-of-the-art methods on MSVD and MSR-VTT datasets respectively. 
We can see that OA-BTG outperforms all the compared methods on popular evaluation metrics, which verifies the effectiveness of bidirectional temporal graph and the object-aware aggregation proposed in our approach.


Among the compared methods, 
STAT \cite{tu2017video} combines spatial and temporal attention, where the spatial attention selects relevant objects while temporal attention selects important frames.
hLSTMat \cite{song2017hierarchical} utilizes an adjusted temporal attention mechanism to distinguish visual words and non-visual words during sentence generation. LSTM-GAN \cite{yang2018video} introduces adversarial learning for video captioning. Different from them, our OA-BTG approach focus on capturing detailed temporal trajectories for objects in video, as well as learning discriminative visual representations. OA-BTG constructs bidirectional temporal graph and performs object-aware feature aggregation to achieve above goals, which helps to generate accurate and fine-grained captions for better performance.

TSA-ED \cite{wu2018interpretable} also utilizes trajectory information, which introduces a trajectory structured attentional encoder-decoder network which explores the fine-grained motion information. Although it extracts dense point trajectories, it loss the object-aware information. While the trajectory extraction in our OA-BTG approach is applied on the object regions, thus captures the object semantics and its temporal dynamics, which play a key role for generating accurate sentence and improve the video captioning performance.


Similarly, the recent work SeqVLAD \cite{xu2018sequential} also incorporates a trainable VLAD process into the sequence learning framework to mine fine motion information in successive video frames. Our OA-BTG approach obtains higher performance for the following two reasons: (1) OA-BTG applies aggregation process on object regions, which can capture the object-aware semantic information. (2) OA-BTG also constructs bidirectional temporal graph to extract the temporal trajectories for each object instance, which captures the detailed temporal dynamics in video content. Thus, OA-BTG achieves better captioning performance.

\subsection{Ablation Study}
In this subsection, we study in detail about the impact of each component of our proposed OA-BTG. The corresponding results are shown in Table \ref{baselines}. The baseline method (denoted as BASELINE) only applies a learnable VLAD model on global frame sequences. The methods in the second row refer to methods with object-aware aggregation with single-directional temporal graph constructed along or reversely along the temporal order. It can be observed that both methods with object-aware aggregation outperform the baseline in popular metrics. For example, ``OA with Backward TG'' improves the performance on BLEU@4, METEOR, CIDEr scores by 0.6\%, 1.3\%, 1.7\% respectively on MSVD dataset, and 1.2\%, 1.0\%, 3.0\% respectively on MSR-VTT dataset. These results verify the effectiveness of object-aware aggregation in our proposed approach.

Comparing OA-BTG with single-directional baseline (OA + Forward/Backward TG), it can be observed that OA-BTG achieves better performance. Taking MSVD dataset for example, OA-BTG obtains the average improvements of 3.25\%, 1.15\%, 5.45\% on BLEU@4, METEOR, CIDEr scores, respectively, which indicates the effectiveness of bidirectional temporal graph. Similarly, improvements can also be found on MSR-VTT dataset. 

Finally, OA-BTG facilitates the baseline method with both innovations of object-aware aggregation and bidirectional temporal graph, and the comparison between OA-BTG and the baseline method definitely verifies the overall effectiveness of our proposed approach.

\begin{figure*}[!t]
    \begin{center}\includegraphics[width=0.95\linewidth]{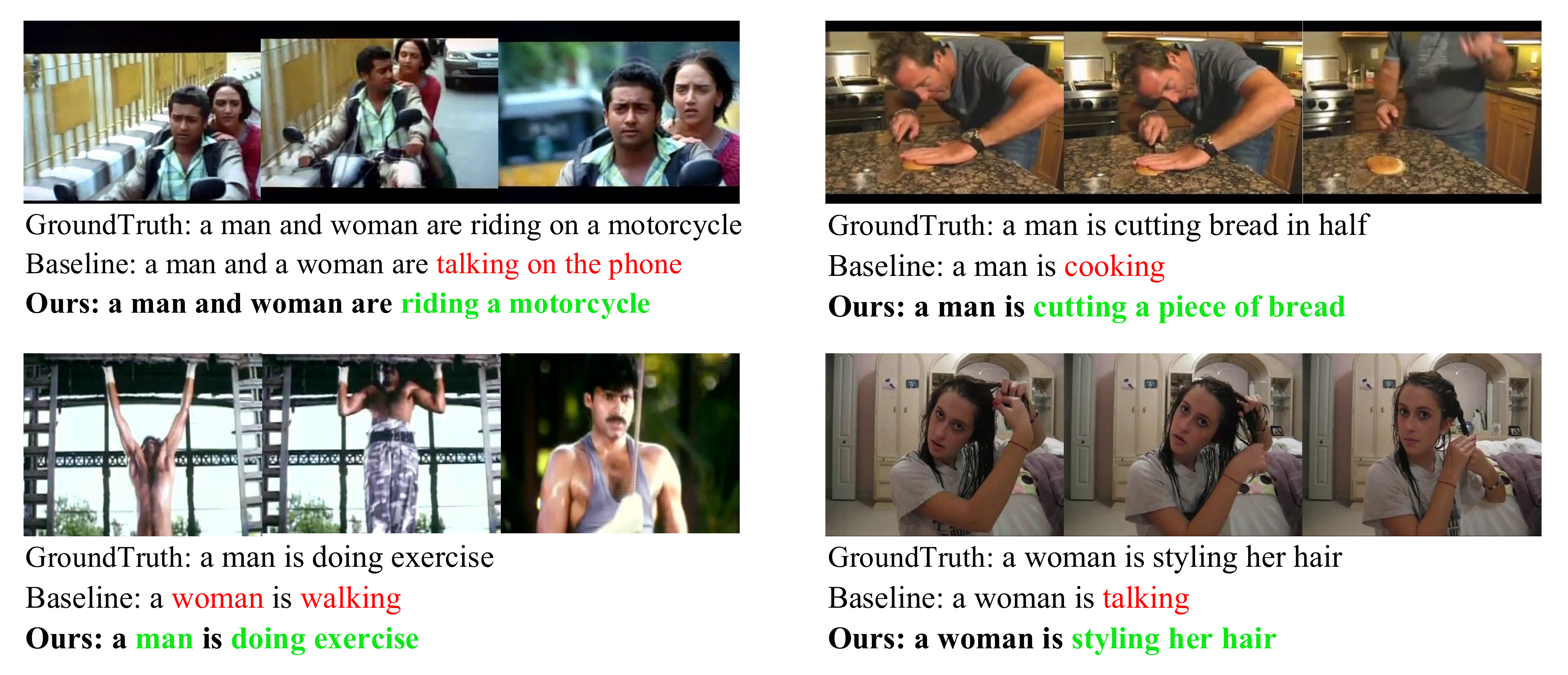}
    \caption{Successful cases of the description examples generated by our OA-BTG approach. Examples of baseline
method are presented for comparison.
    }
    \label{results_success}
    \end{center}
\end{figure*}
\begin{figure*}[!t]
    \begin{center}\includegraphics[width=0.95\linewidth]{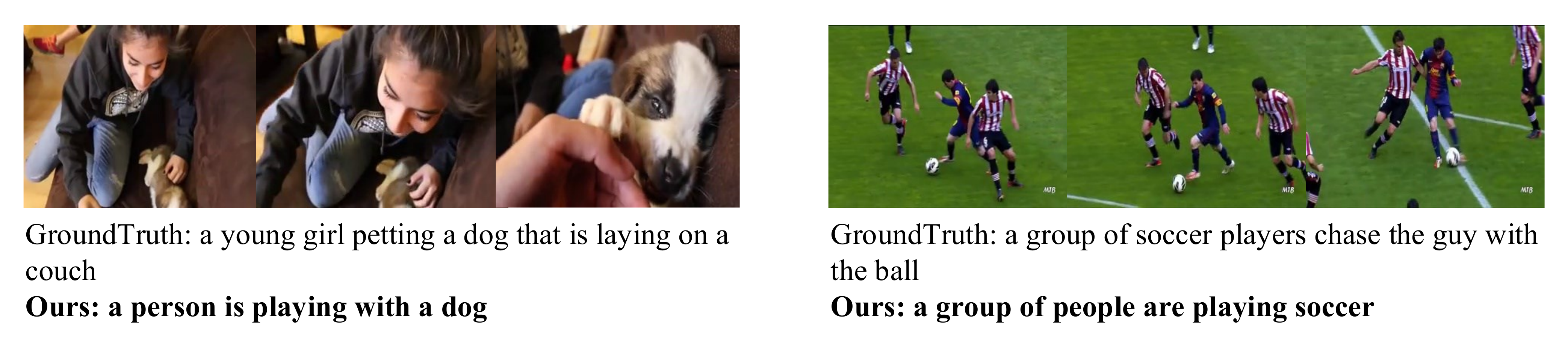}
    \caption{Failure cases of the description examples generated by our OA-BTG approach. 
    }
    \label{results_failure}
    \end{center}
\end{figure*}
\subsection{Qualitative Analysis}
Figures \ref{results_success} and \ref{results_failure} present some successful and failure cases of the generated descriptions by our OA-BTG. From figure \ref{results_success}, it can be seen that our approach indeed improves the video captioning by capturing objects and their detailed temporal information. For instance, the example in top-left demonstrates that our approach can generate accurate depiction of actions by modeling the temporal trajectories of each object. The example in top-right shows that our approach not only expresses the correct semantics, but also is capable of capturing detailed actions so as to generate fine-grained description (``cutting a piece of bread'') rather than a general one (``cooking''). Overall, all these comparisons verify the effectiveness of our proposed method.
Figure \ref{results_failure} shows two failure cases, where our OA-BTG approach fails to describe ``on a couch'' and ``chase''. It needs to not only model salient objects with their trajectories, but also understand interaction relationships among objects, which is very challenging. However, our approach still successfully describes ``playing with a dog'', ``a group of people'', ``playing soccer'' by modeling
object-aware temporal information.
\section{Conclusion}

In this paper, we have proposed a novel video captioning approach based on \emph{object-aware aggregation with bidirectional temporal graph (OA-BTG)}, which captures the detailed temporal dynamics on salient object instances, as well as learns discriminative spatio-temporal representations for complex video content by aggregating fine local information on object-aware regions and frames. First, a bidirectional temporal graph is constructed to capture temporal trajectories for each object instance in two complementary directions, which exploits the detailed temporal dynamics in video for generating accurate and fine-grained captions. Then, object-aware aggregation is performed to encoding the fine spatio-temporal information for salient objects and global context. The integrity of our model, with contributions of the bidirectional temporal graph and object-aware aggregation captures crucial spatial and temporal cues simultaneously, and thus boosting the performance.

In the future, we will explore how to construct more effective graph to model the relations among different object instances, as well as explore their interactions between the backward temporal sequences in an end-to-end model.

\section{Acknowledgments}
This work was supported by the National Natural Science Foundation of China under Grant 61771025 and Grant 61532005.

\newpage
{\small
\bibliographystyle{ieee_fullname}
\bibliography{reference}

\begin{thebibliography}{10}

\bibitem{yao2015describing}
Li~Yao, Atousa Torabi, Kyunghyun Cho, Nicolas Ballas, Christopher Pal, Hugo
  Larochelle, and Aaron Courville.
\newblock Describing videos by exploiting temporal structure.
\newblock In {\em ICCV}, pages 4507--4515, 2015.

\bibitem{yang2017catching}
Ziwei Yang, Yahong Han, and Zheng Wang.
\newblock Catching the temporal regions-of-interest for video captioning.
\newblock In {\em ACM MM}, pages 146--153, 2017.

\bibitem{baraldi2017hierarchical}
Lorenzo Baraldi, Costantino Grana, and Rita Cucchiara.
\newblock Hierarchical boundary-aware neural encoder for video captioning.
\newblock In {\em CVPR}, pages 3185--3194, 2017.

\bibitem{pan2016hierarchical}
Pingbo Pan, Zhongwen Xu, Yi~Yang, Fei Wu, and Yueting Zhuang.
\newblock Hierarchical recurrent neural encoder for video representation with
  application to captioning.
\newblock In {\em CVPR}, pages 1029--1038, 2016.

\bibitem{arandjelovic2016netvlad}
Relja Arandjelovic, Petr Gronat, Akihiko Torii, Tomas Pajdla, and Josef Sivic.
\newblock Netvlad: Cnn architecture for weakly supervised place recognition.
\newblock In {\em CVPR}, pages 5297--5307, 2016.

\bibitem{xu2018sequential}
Youjiang Xu, Yahong Han, Richang Hong, and Qi~Tian.
\newblock Sequential video vlad: Training the aggregation locally and
  temporally.
\newblock {\em IEEE Transactions on Image Processing (TIP)}, 27(10):4933--4944,
  2018.

\bibitem{guadarrama2013youtube2text}
Sergio Guadarrama, Niveda Krishnamoorthy, Girish Malkarnenkar, Subhashini
  Venugopalan, Raymond Mooney, Trevor Darrell, and Kate Saenko.
\newblock Youtube2text: Recognizing and describing arbitrary activities using
  semantic hierarchies and zero-shot recognition.
\newblock In {\em ICCV}, pages 2712--2719, 2013.

\bibitem{rohrbach2013translating}
Marcus Rohrbach, Wei Qiu, Ivan Titov, Stefan Thater, Manfred Pinkal, and Bernt
  Schiele.
\newblock Translating video content to natural language descriptions.
\newblock In {\em ICCV}, pages 433--440, 2013.

\bibitem{krishnamoorthy2013generating}
Niveda Krishnamoorthy, Girish Malkarnenkar, Raymond~J Mooney, Kate Saenko, and
  Sergio Guadarrama.
\newblock Generating natural-language video descriptions using text-mined
  knowledge.
\newblock In {\em AAAI}, pages 541--547, 2013.

\bibitem{bahdanau2014neural}
Dzmitry Bahdanau, Kyunghyun Cho, and Yoshua Bengio.
\newblock Neural machine translation by jointly learning to align and
  translate.
\newblock In {\em ICLR}, pages 1--15, 2015.

\bibitem{venugopalan2015sequence}
Subhashini Venugopalan, Marcus Rohrbach, Jeffrey Donahue, Raymond Mooney,
  Trevor Darrell, and Kate Saenko.
\newblock Sequence to sequence-video to text.
\newblock In {\em ICCV}, pages 4534--4542, 2015.

\bibitem{wang2018m3}
Junbo Wang, Wei Wang, Yan Huang, Liang Wang, and Tieniu Tan.
\newblock M3: Multimodal memory modelling for video captioning.
\newblock In {\em CVPR}, pages 7512--7520, 2018.

\bibitem{ZhangP19Hierachical}
Junchao Zhang and Yuxin Peng.
\newblock Hierarchical vision-language alignment for video captioning.
\newblock In {\em MMM}, pages 42--54, 2019.

\bibitem{venugopalan2014translating}
Subhashini Venugopalan, Huijuan Xu, Jeff Donahue, Marcus Rohrbach, Raymond
  Mooney, and Kate Saenko.
\newblock Translating videos to natural language using deep recurrent neural
  networks.
\newblock In {\em ACL}, pages 1494--1504, 2015.

\bibitem{chen2017video}
Shizhe Chen, Jia Chen, Qin Jin, and Alexander Hauptmann.
\newblock Video captioning with guidance of multimodal latent topics.
\newblock In {\em ACM MM}, pages 1838--1846, 2017.

\bibitem{xu2017learning}
Jun Xu, Ting Yao, Yongdong Zhang, and Tao Mei.
\newblock Learning multimodal attention lstm networks for video captioning.
\newblock In {\em ACM MM}, pages 537--545, 2017.

\bibitem{hori2017attention}
Chiori Hori, Takaaki Hori, Teng-Yok Lee, Ziming Zhang, Bret Harsham, John~R
  Hershey, Tim~K Marks, and Kazuhiko Sumi.
\newblock Attention-based multimodal fusion for video description.
\newblock In {\em ICCV}, pages 4203--4212, 2017.

\bibitem{WangWW18}
Xin Wang, Wang Yuan-Fang, and William~Yang Wang.
\newblock Watch, listen, and describe: Globally and locally aligned cross-modal
  attentions for video captioning.
\newblock In {\em ACL}, pages 795--801, 2018.

\bibitem{tran2015learning}
Du~Tran, Lubomir Bourdev, Rob Fergus, Lorenzo Torresani, and Manohar Paluri.
\newblock Learning spatiotemporal features with 3d convolutional networks.
\newblock In {\em ICCV}, pages 4489--4497, 2015.

\bibitem{xu2013feature}
Zhongwen Xu, Yi~Yang, Ivor Tsang, Nicu Sebe, and Alexander~G Hauptmann.
\newblock Feature weighting via optimal thresholding for video analysis.
\newblock In {\em ICCV}, pages 3440--3447, 2013.

\bibitem{chen2011collecting}
David~L Chen and William~B Dolan.
\newblock Collecting highly parallel data for paraphrase evaluation.
\newblock In {\em ACL}, pages 190--200, 2011.

\bibitem{xu2016msr}
Jun Xu, Tao Mei, Ting Yao, and Yong Rui.
\newblock Msr-vtt: A large video description dataset for bridging video and
  language.
\newblock In {\em CVPR}, pages 5288--5296, 2016.

\bibitem{yang2018video}
Yang Yang, Jie Zhou, Jiangbo Ai, Yi~Bin, Alan Hanjalic, Heng~Tao Shen, and
  Yanli Ji.
\newblock Video captioning by adversarial lstm.
\newblock {\em IEEE Transactions on Image Processing (TIP)}, 27(11):5600--5611,
  2018.

\bibitem{papineni2002bleu}
Kishore Papineni, Salim Roukos, Todd Ward, and Wei-Jing Zhu.
\newblock Bleu: a method for automatic evaluation of machine translation.
\newblock In {\em ACL}, pages 311--318, 2002.

\bibitem{banerjee2005meteor}
Satanjeev Banerjee and Alon Lavie.
\newblock Meteor: An automatic metric for mt evaluation with improved
  correlation with human judgments.
\newblock In {\em Proceedings of the acl workshop on intrinsic and extrinsic
  evaluation measures for machine translation and/or summarization}, pages
  65--72, 2005.

\bibitem{vedantam2015cider}
Ramakrishna Vedantam, C~Lawrence~Zitnick, and Devi Parikh.
\newblock Cider: Consensus-based image description evaluation.
\newblock In {\em CVPR}, pages 4566--4575, 2015.

\bibitem{chen2015microsoft}
Xinlei Chen, Hao Fang, Tsung-Yi Lin, Ramakrishna Vedantam, Saurabh Gupta, Piotr
  Doll{\'a}r, and C~Lawrence Zitnick.
\newblock Microsoft coco captions: Data collection and evaluation server.
\newblock {\em arXiv preprint arXiv:1504.00325}, 2015.

\bibitem{he2017mask}
Kaiming He, Georgia Gkioxari, Piotr Doll{\'a}r, and Ross Girshick.
\newblock Mask r-cnn.
\newblock In {\em ICCV}, pages 2980--2988. IEEE, 2017.

\bibitem{he2016deep}
Kaiming He, Xiangyu Zhang, Shaoqing Ren, and Jian Sun.
\newblock Deep residual learning for image recognition.
\newblock In {\em CVPR}, pages 770--778, 2016.

\bibitem{lin2014microsoft}
Tsung-Yi Lin, Michael Maire, Serge Belongie, James Hays, Pietro Perona, Deva
  Ramanan, Piotr Doll{\'a}r, and C~Lawrence Zitnick.
\newblock Microsoft coco: Common objects in context.
\newblock In {\em ECCV}, pages 740--755. Springer, 2014.

\bibitem{song2018deterministic}
Jingkuan Song, Yuyu Guo, Lianli Gao, Xuelong Li, Alan Hanjalic, and Heng~Tao
  Shen.
\newblock From deterministic to generative: Multimodal stochastic rnns for
  video captioning.
\newblock {\em IEEE transactions on neural networks and learning systems
  (TNNLS)}, 2018.

\bibitem{wu2018multi}
Aming Wu and Yahong Han.
\newblock Multi-modal circulant fusion for video-to-language and backward.
\newblock In {\em IJCAI}, pages 1029--1035, 2018.

\bibitem{wang2018reconstruction}
Bairui Wang, Lin Ma, Wei Zhang, and Wei Liu.
\newblock Reconstruction network for video captioning.
\newblock In {\em CVPR}, pages 7622--7631, 2018.

\bibitem{wu2018interpretable}
Xian Wu, Guanbin Li, Qingxing Cao, Qingge Ji, and Liang Lin.
\newblock Interpretable video captioning via trajectory structured
  localization.
\newblock In {\em CVPR}, pages 6829--6837, 2018.

\bibitem{gao2017video}
Lianli Gao, Zhao Guo, Hanwang Zhang, Xing Xu, and Heng~Tao Shen.
\newblock Video captioning with attention-based lstm and semantic consistency.
\newblock {\em IEEE Transactions on Multimedia (TMM)}, 19(9):2045--2055, 2017.

\bibitem{tu2017video}
Yunbin Tu, Xishan Zhang, Bingtao Liu, and Chenggang Yan.
\newblock Video description with spatial-temporal attention.
\newblock In {\em ACM MM}, pages 1014--1022. ACM, 2017.

\bibitem{song2017hierarchical}
Jingkuan Song, Lianli Gao, Zhao Guo, Wu~Liu, Dongxiang Zhang, and Heng~Tao
  Shen.
\newblock Hierarchical lstm with adjusted temporal attention for video
  captioning.
\newblock In {\em IJCAI}, pages 2737--2743, 2017.

\bibitem{zhu2017bidirectional}
Linchao Zhu, Zhongwen Xu, and Yi~Yang.
\newblock Bidirectional multirate reconstruction for temporal modeling in
  videos.
\newblock In {\em CVPR}, pages 2653--2662, 2017.

\bibitem{zhang122017task}
Xishan Zhang, Ke~Gao, Yongdong Zhang, Dongming Zhang, Jintao Li, and Qi~Tian.
\newblock Task-driven dynamic fusion: Reducing ambiguity in video description.
\newblock In {\em CVPR}, pages 6250--6258, 2017.

\end{thebibliography}
}

\end{document}